\documentclass[10pt,twocolumn,letterpaper]{article}

\usepackage{iccv}
\usepackage{times}
\usepackage{epsfig}
\usepackage{graphicx}
\usepackage{amsmath}
\usepackage{amssymb}
\usepackage{bbm}
\usepackage{bm}

\usepackage{subfigure}
\usepackage{multirow}
\usepackage{threeparttable}
\usepackage{makecell}
\usepackage{bbding}
\usepackage[percent]{overpic}
\usepackage[ruled,linesnumbered]{algorithm2e}

\usepackage[pagebackref=true,breaklinks=true,colorlinks,bookmarks=false]{hyperref}

\iccvfinalcopy 

\def\iccvPaperID{7204} 

\ificcvfinal\fi

\begin{document}

\title{\textit{Self}-EMD: Self-Supervised Object Detection without ImageNet}

\author{Songtao Liu,\ \ \  Zeming Li,\ \ \  Jian Sun\\
	MEGVII Technology\\
	{\tt\small \{liusongtao, lizeming, sunjian\}@megvii.com}
}

\maketitle
\ificcvfinal\thispagestyle{empty}\fi
	
\begin{abstract}
	In this paper, we propose a novel self-supervised representation learning method, Self-\emph{EMD}, for object detection. Our method directly trained on unlabeled non-iconic image datasets like COCO, instead of commonly used iconic-object image datasets like ImageNet. We keep the convolutional feature maps as the image embedding to preserve spatial structures and adopt Earth Mover's Distance (EMD) to compute the similarity between two embeddings. Our Faster R-CNN (ResNet50-FPN) baseline achieves 39.8\% mAP on COCO, which is on par with the state-of-the-art self-supervised methods pre-trained on ImageNet. More importantly, it can be further improved to 40.4\% mAP with more unlabeled images, showing its great potential for leveraging more easily obtained unlabeled data. Code will be made available.         
\end{abstract}
	
\section{Introduction}

Unsupervised visual representation learning has attracted considerable attention, which aims to generate better feature representation with large amounts of unlabeled data. 
Recent self-supervised learning methods~\cite{examplar,memorybank,moco,mocov2,byol,swav} have achieved comparable or better results than the supervised counterpart. 
They train CNN models on the unlabeled ImageNet dataset by performing instance-level classification task, which maximizes agreement between differently transformed views of the same image and optionally minimizes agreement between views of different images. 
	
\begin{figure}[t]
	\begin{center}
		\includegraphics[width=0.99\linewidth]{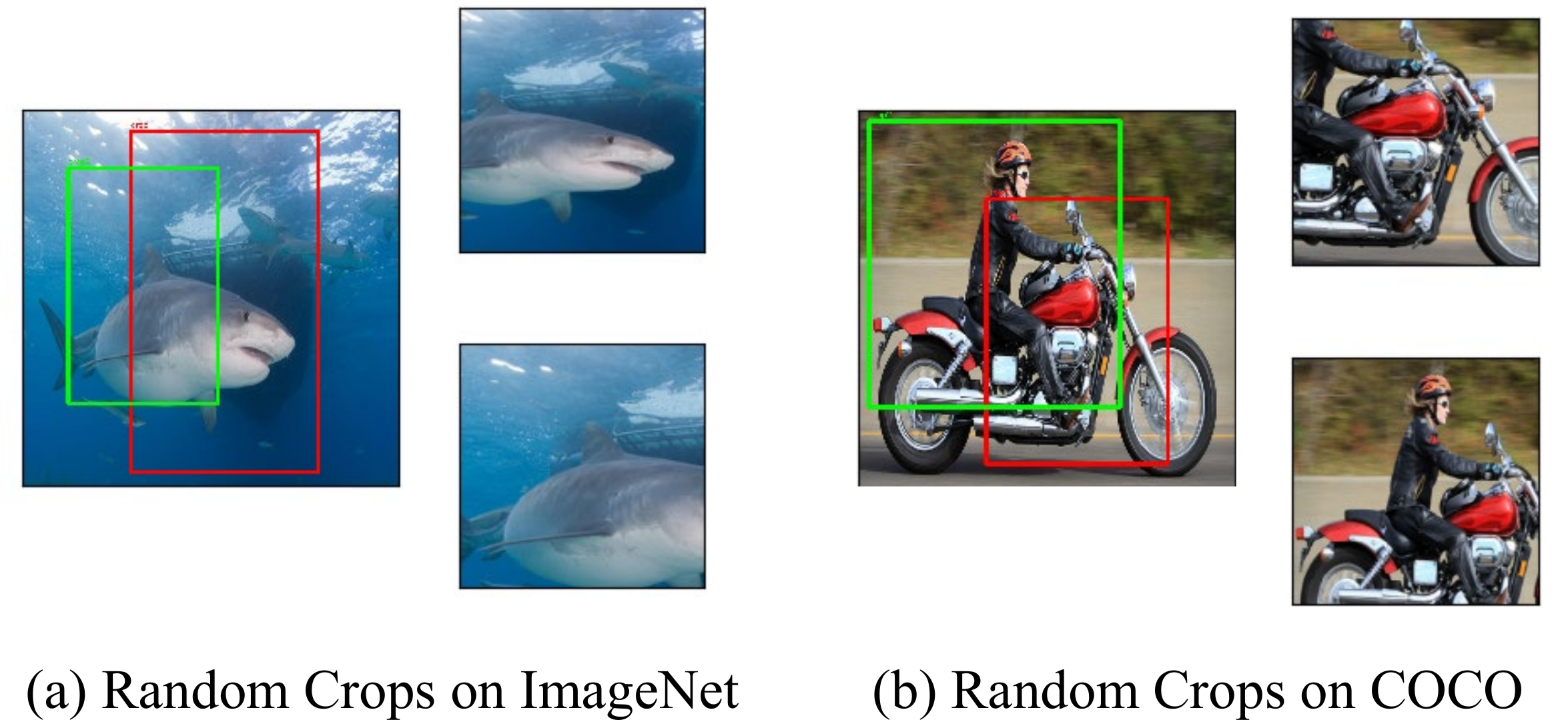}
	\end{center}
	\caption{Visualization of different random cropped views on ImageNet and COCO dataset. On ImageNet, as the images are pre-cropped to be \textit{object-centric}, different crops of the same image are from the same object. But on COCO, where each image contains multi-objects, different crops might correspond different objects. This inconsistent noise may hurt the effectiveness of self-supervised learning methods.}
	\label{fig:crop}
\end{figure}

The latent prior of self-supervised learning pipeline is that different views/crops of the same image correspond to the same object, as shown in Fig.~\ref{fig:crop} (a). So maximizing their agreement can learn useful features.
This critical prior in fact highly relies on the underlying biases of the pre-training dataset: ImageNet is an \textit{object-centric} dataset that ensures the latent prior. 
When the extra effort of collecting and cleaning data is taken into account, the unlabeled ImageNet dataset is in fact not free. 
On the other hand, non-iconic images are easy to collect but may contain multi-objects, such as COCO~\cite{MSCOCO} dataset. As illustrated in Fig.~\ref{fig:crop} (b), different random crops may correspond to different objects, where the effectiveness of the self-supervised methods is open to doubt.  
Moreover, the distinctive representation learned from instance-level classification task may be not the suitable feature for object detection. As it applies a global pooling layer to generate vector embeddings, it may destroy image spatial structures and lose local information, while detectors need to be sensitive to spatially localization.
	
\begin{figure}[t]
	\begin{center}
		\includegraphics[width=1.03\linewidth]{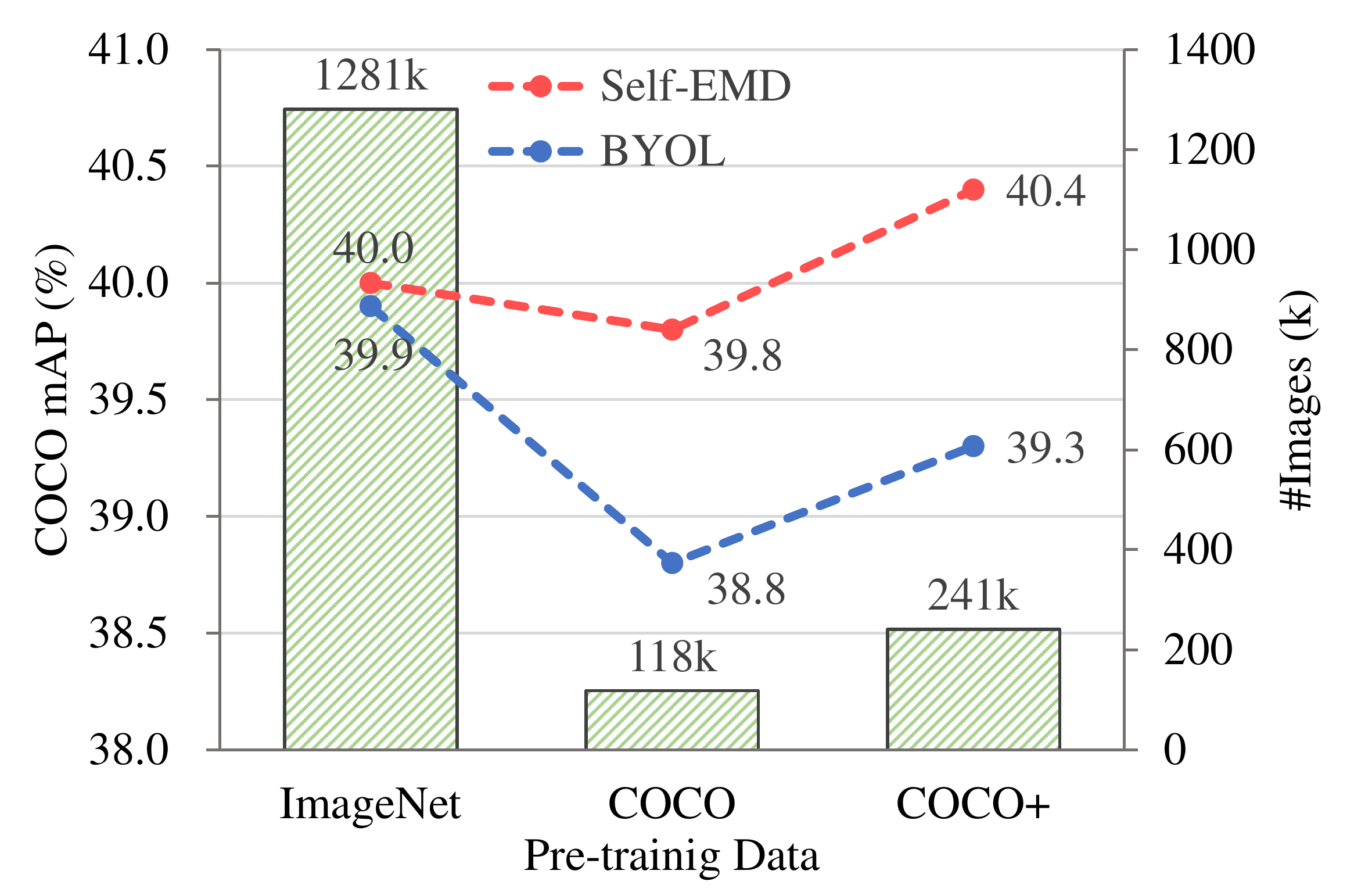}
	\end{center}
	\caption{Comparisons of pre-trained models by fine-tuning on COCO. `COCO+' denotes the COCO \textit{train} 2017 set plus the COCO \textit{unlabel} set. The results are fine-tuned on COCO \textit{train} 2017 for 90k iterations and evaluated on COCO \textit{val} 2017. All results are based on the same Faster R-CNN with a ResNet50 backbone and a FPN neck. }
	\label{fig:0}
\end{figure}
In this paper, 
we introduce a novel self-supervised pre-training method which directly trained on non-iconic image dataset like COCO (unlabelled) to obtain better feature representation for object detection. 
Instead of using global pooling, we keep the convolutional feature maps as the image embeddings, preserving local and spatial information. This dense representation introduces a new difficulty that we need a desirable metric to measure the similarity between two feature maps. As the corresponding two crops may have different objects at different spatial locations in a non-iconic image, the metric should pick up the optimal matching among local patches without correspondence supervision and minimize the noise from the irrelevant regions. 
	
To address this issue, we employ the discrete Earth Mover's Distance (EMD) to compute the spatial similarity among all location pairs, named as \textit{Self}-EMD. EMD is the metric for computing distance between structural representations. Given the distance between all element pairs from two sets, EMD acquires the optimal matching flows between two structures that have the minimum cost. We define the cost between any two locations across two feature maps as the cosine similarity of the local embeddings, and design a proper marginal weights for EMD constraints. Such discrete EMD has the formulation of the transportation problem and can be quickly solved by Sinkhorn-Knopp iteration~\cite{sinkhorn,sinkhorn2}.         
	
Our thorough experiments on PASCAL VOC~\cite{Pascal-voc} and MS COCO~\cite{MSCOCO} datasets with several typical detectors are conducted to demonstrate the effectiveness of our method. 
As shown in Fig.~\ref{fig:0}, after pre-training on the unlabeled COCO training set (118K images), the proposed \textit{Self}-EMD get an mAP of 39.8\% with a standard Res50-FPN Faster R-CNN~\cite{FPN} detector on COCO fine-tuning task, surpassing both the state-of-the-art self-supervised method BYOL~\cite{byol} (38.8\%) and the ImageNet (1.28M images) supervised baseline (39.1\%). With pre-training on an augmented ``COCO+" dataset (COCO training set and unlabeled COCO set, 241K in total), \textit{Self}-EMD can further improve the detector to 40.4\% mAP, showing its merit of benefiting from more unlabeled data.   

\section{Related Work}
\paragraph{Generic Object Detection.} Most modern object detectors, such as Faster R-CNN \cite{FasterCNN}, Mask R-CNN \cite{mask-rcnn} and RetinaNet \cite{focal-loss}, employ a ``pre-training and fine-tuning'' paradigm that pre-train the networks for ImageNet classification \cite{imagenet} and then transfer the parameters for detection fine-tuning. Recently, \cite{cheaper} switch the pre-training to the target detection domain with a montage manner to improve the efficiency. These works demonstrate significant benefit of learning from large-scale data, but they also suffer from the high cost of labeling data for real world application. 
	
Given the success of transfer learning paradigm, later works \cite{dsod, pretrain} demonstrate that it is often possible to match fine-tuning accuracy when training from scratch for object detection task. However, \cite{pretrain} also verified that the performance of this paradigm significantly drops when it enter a small data regime. Since the expense of labeling detection data is much higher than classification, the reliance of large-scale labeled data still limits the application of modern object detectors.
	
\paragraph{Self Supervised Learning.} Many self-supervised methods~\cite{self1,self2,self3,self4,self5,self6,self7,self8,self9,self10} manipulate the input data to extract a supervised signal in the form of a pretext task. 
Recently, an instance-level classification task has achieved promising results in this field. Instance-level classification considers each image in a dataset as its own class, which is firstly proposed in \cite{examplar}.
As this approach is quite intractable for large-scale datasets, \cite{memorybank} mitigate this issue by replacing the classifier with a memory bank that stores previously-computed representations. They rely on noise contrastive estimation to compare instances, which is a special form of contrastive learning. MoCo~\cite{moco,mocov2} improve the training of contrastive methods by storing representations from a momentum encoder instead of the trained
network. SimCLR~\cite{simclr} show that the memory bank can be entirely replaced with the elements from the same batch if the batch is large enough. More recently, BYOL~\cite{byol} and SWAV~\cite{swav} avoid comparing every pair of images, especially for negative pairs. BYOL directly bootstrap the representations by attracting the different features from the same instance, while SWAV maps the image features to a set of trainable prototype vectors.
	
The success of instance-level classification actually relies on the underlying biases of ImageNet: each image is object-centric to ensures different views and crops of the same image correspond to the same object. If we take into account the extra effort of collecting and cleaning data, the self-supervised ``pre-training'' step is in fact still not free.   
	
\begin{figure*}[!t]
		\begin{center}
			\includegraphics[width=0.90\linewidth]{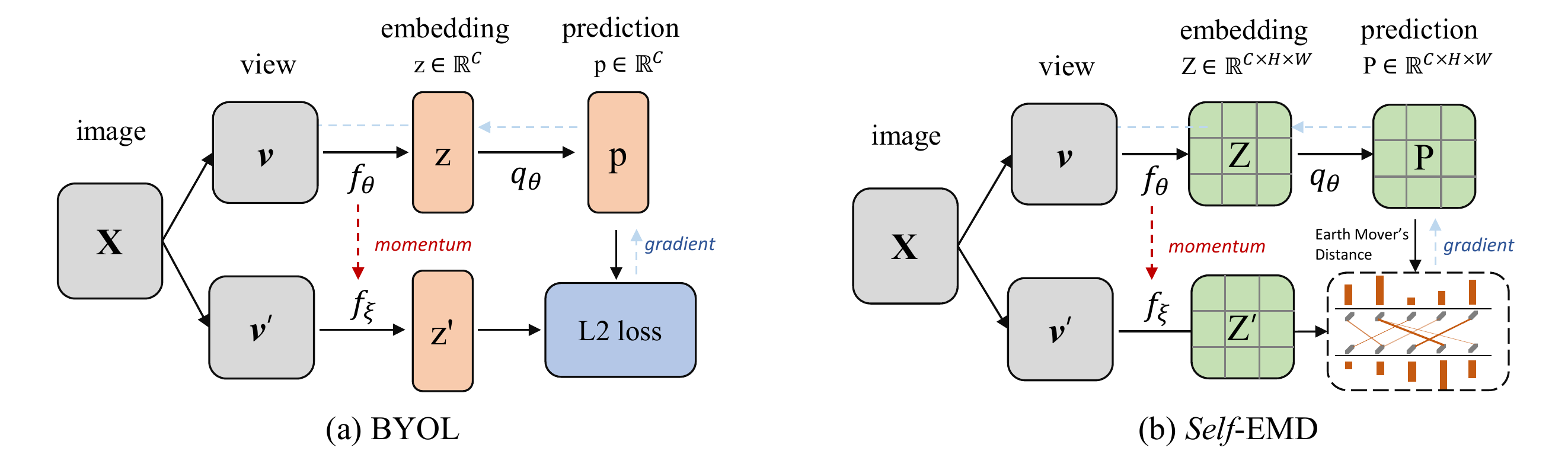}
		\end{center}
		\vspace{-0.5cm}
		\caption{The framework of self-supervised methods. (a) original BYOL: there are two global pooling layers behind $f_\theta$ and $f_\xi$ respectively to generate vector embeddings, which are easy to learn global semantic features but lose spatial information for detectors; (b) the proposed \textit{Self}-EMD: we drop the global pooling layers to keep the spatial representations of the crops and conduct the EMD metric to calculate the similarity between two image representations.}
		\label{fig:pipeline}
\end{figure*}
	\paragraph{Earth Mover's Distance.} Earth Mover's Distance has been already adopted in many previous computer vision tasks. It serves as a metric for color and texture based image retrieval in \cite{emd-retrieval}. Moreover, it is also investigated in visual tracking problem \cite{emd-tracking1,emd-tracking2,emd-tracking3}. Recently, \cite{deepemd} uses EMD to compute a structural distance between image representations in the few-shot classifications task, which shares some similarity with this paper. However, EMD in \cite{deepemd} and the other above works is conducted in the supervised manner, which contains rich semantic information for computing the distance. In this paper, we delicately design the components of \textit{Self}-EMD to conduct the first successful application of EMD in self-supervised learning.

\section{Method}
	We start by briefly introducing BYOL~\cite{byol} as our baseline. 
	Many successful self-supervised learning approaches build upon the cross-view prediction framework. Typically, these approaches learn representations by predicting different views (\emph{e.g.}, different random crops and transforms, denoted as $v,v'$) of the same image from one another. This framework relies on the object-centric bias in ImageNet to ensure that different views correspond to the same object.
	Among them, the recent BYOL achieves the state-of-the-art performance, we thus choose it as our baseline. Then, we introduce how to apply EMD to learn spatial feature representations for object detection and other pre-training tricks.
	
	\subsection{Description of BYOL}
	BYOL's goal is to learn a representation by maximizing the similarity between two different views from the same image. As shown in Fig.~\ref{fig:pipeline} (a), BYOL uses two networks to extract vector embeddings from two views separately. The first view $v$ is fed into the query encode network which consists of a backbone $f_\theta$ and a predictor $q_\theta$. The second view $v'$ is extracted by the key encoder that has the same architecture as the query encoder except the predictor. Moreover, the weights of the key encoder $\xi$ are an exponential moving average of the query parameters $\theta$. Given a momentum rate $m \in [0,1]$, $\xi$ is updated by the following rule,
	\begin{equation}
	\xi \leftarrow m\xi + (1-m)\theta.
	\end{equation}
	For the query parameters $\theta$, BYOL minimizes a similarity loss between two different views to update it. The loss is defined as: 
	\begin{equation}
	\label{byol}
	\mathcal{L}_{BYOL} \triangleq\|p-z'\|_{2}^{2}=2-2 \cdot \frac{\left\langle 
		p, z'\right\rangle}{\|p\|_{2} \cdot\|z'\|_{2}},
	\end{equation}
	where $p$ and $z'$ are the embeddings from the query encoder and key encoder respectively, $\left\langle\cdot \right\rangle$ denotes the inner product. The symmetric loss of swapping $v$ and $v'$ is also computed. 
	
	At the end of training, only the backbone in the query encoder is kept for representation and downstream tasks.     
	
	\subsection{EMD for Feature Maps}
	\label{sec:3.2}
	
	The cross-view framework of BYOL hypotheses that the two different views are from the same object, but this bias is only ensured on the object-centric images. When training on the unlabeled multi-object images, the noise from multi-objects may impede the learning process, leading to inferior performance of the detection fine-tuning task. Moreover, BYOL and other self-supervised learning methods adopt a global pooling layer to generate vector embeddings, which ruins spatial structures and local information.    
	
	To alleviate the reliance of object-centric bias and the restraint of global pooling, we take away the pooling layer and replace the MLP head with convolution layers to keep the spatial feature maps $\textbf{F} \in \mathcal{R}^{H\times W\times C}$, where $H$ and $W$ denote the spatial size of the feature map and $C$ is the number of channels, as shown in Fig.~\ref{fig:pipeline} (b).
	
	\paragraph{Earth Mover's Distance for Feature Maps.} Now the problem lies in an appropriate metric to measure the similarity or distance between the feature map of two crops. As the two crops are from one non-iconic image, they may contain the same or different objects at different locations. This makes the L2 distance in BYOL wrecked since we have no annotation to spatially align object features of those two crops. In this work, we adopt Earth Mover's Distance (EMD) to automatically find the corresponding maps and then calculate the distance.           
	
	Earth Mover's Distance~\cite{emd-retrieval,deepemd} is a distance measure between two sets of weighted objects or distributions, which is built up the basic distance between individual objects. Its discrete form can be formalized to the well-studied optimal transportation problem (OTP). Specially, for two feature maps $\textbf{X}, \textbf{Y} \in \mathcal{R}^{H\times W\times C}$, we first flatten them into two set of local representations $\mathcal{X} = \{\textbf{x}_i| i=1,2,...HW\}$ and $\mathcal{Y} = \{\textbf{y}_j| j=1,2,...HW\}$, where $\textbf{x}_i$ and $\textbf{y}_j$ denote the local vectors in the corresponding position of feature maps. Then we define the EMD between two feature maps as the minimum ``transport cost'' from ``suppliers'' $\mathcal{X}$ to ``demanders'' $\mathcal{Y}$. Suppose for each supplier $\textbf{x}_i$, it has totally $r_i$ units to transport, and for each demander $\textbf{y}_j$, it requires $c_j$ units. The overall ``transport polytope'' can be formulated as:
	\begin{equation}
	U(r, c) := \{\pi \in \mathcal{R}^{HW\times HW}_+ | \pi\mathbbm{1}=\textbf{r}, \pi^T\mathbbm{1}=\textbf{c}\}.
	\end{equation} 
	Here $\mathbbm{1}$ are vectors of all ones of the appropriate dimension (in our case $\mathbbm{1} \in \mathcal{R}^{HW}$). $\textbf{r}$ and $\textbf{c}$ are vectorized representation of $\{r_i\}$ and $\{c_j\}$, which are also called marginal weights of matrix $\pi$ onto its rows and columns, respectively. 
	
	We then define the cost per unit transported from supplier feature node $\textbf{x}_i$ to demander node $\textbf{y}_j$ as:
	\begin{equation}
	M_{ij} = 1-\frac{\textbf{x}_i^T \textbf{y}_j}{\|\textbf{x}_i\| \|\textbf{y}_j\|},
	\end{equation}     
	where nodes with similar representations tend to generate fewer transport cost between each other. With this notation, we can define the EMD as the optimal transportation problem as:
	\begin{equation}
	d_{C}(r,c):= \min_{\pi \in U(r,c)} \langle\pi, M\rangle,
	\end{equation} 
	where $M$ is the cost matrix of $\{M_{ij}\}$ and $\langle \cdot, \cdot \rangle$ stands for the Frobenius dot-product between two matrices.
	
	This is a linear program which can be solved in polynomial time. For image features, however, the resulting linear program is large, involving the square of feature dimensions and the large batch sizes. We thus address this issue by using a fast iterative solution named \textit{Sinkhorn-Knopp algorithm}~\cite{sinkhorn2}, which introduces a regularization term $E$:
	\begin{alignat}{2}
	\label{eq:6}
	\min_{\pi\in U(r,c)}\langle\pi, M\rangle + \frac{1}{\lambda} E(\pi),
	\end{alignat}
	where $E(\pi)=\pi (\log \pi-1)$ and $\lambda$ is a constant hyper-parameter that controls the intensity of regularization term. The advantage of this regularization term is that the optimal transport plan $\pi^*$ of Eq. (\ref{eq:6}) can be written as: 
	\begin{equation}
	\label{eq:7}
	\begin{split}
	\pi^* = diag(v) P diag(u),
	\end{split}
	\end{equation}
	where $P=e^{-\lambda M}$ is the element-wise exponential of $-\lambda M$ and $v$ and $u$ are two non-negative vectors of scaling coefficients chosen so that the resulting matrix $\pi \in U(r,c)$ (see \cite{sinkhorn2} for a derivation). The vector $v$ and $u$ can be obtained via a simple iteration as following:  
	\begin{alignat}{2}
	&\forall i, \quad v_i^{t+1}\leftarrow\frac{r_i}{\sum_jP_{ij}u_j^{t}} \\
	&\forall j, \quad u_j^{t+1}\leftarrow\frac{c_j}{\sum_iP_{ij}v_i^{t+1}}.
	\end{alignat}	
	
	After repeating this iteration $T$ times ($T=10$ in our case), the approximate optimal plan $\pi^*$ can be obtained by Eq. (\ref{eq:7}). In our case, the high values of $\pi_{ij}^*$ denotes the transport cost from $\textbf{x}_i$ to $\textbf{y}_j$ is small so that we transport units as much as possible, which also indicates that $\textbf{x}_i$ and $\textbf{y}_j$ have similar features so that they may correspond to the same object. Thus the optimal $\pi^*$ actually represents the best corresponding map between the local feature vectors of the two crops. Then we can compute the similarity score $S$ between two image feature representations with:
	\begin{equation}
	S(\textbf{X},\textbf{Y})=\langle\pi, 1-M\rangle,
	\end{equation}
	as $1-M$ denotes the normalized similarity between two local vector features. Then, similar to Eq.~(\ref{byol}), the corresponding loss for two feature maps is defined as:
	\begin{equation}
	\mathcal{L}_{EMD} \triangleq\|\textbf{X}-\textbf{Y}\|_{EMD}=2-2 \cdot S(\textbf{X},\textbf{Y}).
	\end{equation} 
	
	\paragraph{Marginal weights.} The marginal weight of each node (\emph{e.g.}, $r_i$ and $c_j$) plays an important role in the EMD problem, which controls the total supplying units $\sum_{j=1}^{HW}\pi_{ij}$ and demanding units $\sum_{i=1}^{HW}\pi_{ij}$, respectively. Intuitively, the node with a larger weight is more important in the comparison of two sets, and vice versa. In the pioneering works~\cite{deepemd} that adopt EMD for few-shot classification, the weight of each node is set as the dot product between a node feature and the average node feature in the other set, which is denoted as:
	\begin{equation}
	\label{eq:weights}
	r_i=\max\{ \textbf{x}_i^T \cdot \frac{\sum_{j=1}^{HW}\textbf{y}_j}{HW}, 0\}.
	\end{equation}
	It makes sense as for supervised classification, features often have high-level semantic meanings, large weight should be given to the co-occurrent regions while the high-variance background should have less weight. This property seems also be suitable for BYOL that compares two random crops from one image. However, as no semantic label is available for unsupervised training, the features may contain less high-level semantic information. For example, instead of being high-variance in supervised learning, the background features may collapse to several simple patterns, generating high similarity and large marginal weights with Eq.~(\ref{eq:weights}).
	
	To address this issue, we keep the original two embedding vectors in BYOL and the L2 loss in Eq. (\ref{byol}) in our model as they are learned to represent the same image and contain minimum background noise. Then, the marginal weight of each node is transferred as:
	\begin{equation}
	\label{margin:weights}
	r_i=\max\{ \textbf{x}_i^T \cdot \textbf{v}_y, 0\},
	\end{equation}         
	where $\textbf{v}_y$ is the corresponding vector features of $\textbf{Y}$.
	
	\paragraph{Pseudo code for EMD.} Finally, we summarize the overall computation of EMD loss $\mathcal{L}_{EMD}$ as in Algorithm \ref{algo:1} to provide a more clear description. 	
	  
	\begin{algorithm}
		\caption{The procedure of EMD loss}
		\label{algo:1}
		\KwData{$\textbf{X},\textbf{Y} \in \mathcal{R}^{H\times W\times C}$, $\textbf{v}_x, \textbf{v}_y \in \mathcal{R}^{C \times 1} $}
		\KwResult{$\mathcal{L}_{EMD}$}
		$M_{ij} \leftarrow 1-\frac{\textbf{x}_i^T \textbf{y}_j}{\|\textbf{x}_i\| \|\textbf{y}_j\|}$ \;
		$r_i\leftarrow\max\{ \textbf{x}_i^T \cdot \textbf{v}_y, 0\}, c_j\leftarrow\max\{ \textbf{y}_j^T \cdot \textbf{v}_x, 0\}$ \;
		Computing the optimal transport $\pi^*\leftarrow diag(v) M diag(u)$ with Sinkhorn iteration \; 
		$S(\textbf{X},\textbf{Y})\leftarrow \langle\pi, 1-M\rangle$ \;
		$\mathcal{L}_{EMD} \leftarrow 2-2 \cdot S(\textbf{X},\textbf{Y})$ \;
		
	\end{algorithm}
	
	\subsection{Scale Invariant Training}
	\label{sec:3.3}
	
	Another gap between the supervised learning method and object detection is about scale invariance, as object detection involves not only classification but also localization, where the threat of varying scales of objects is particularly
	evident. The scale invariance has been widely explored in generic supervised detectors, while it remains largely intact in self-supervised learning. In fact, the self-supervised methods like BYOL implicitly learn scale-invariant representations in some degree through the random scale transforms among different views. However, as the scale of objects varies a lot, some explicit scale invariant learning designs are needed.   
	
	One common practical training trick in supervised object detection is multi-scale training, where the input images are resized to multi scales and the annotation is modified respectively. In self-supervised task, however, we have no label to accommodate for training consistency. We thus introduce a third transformed view $v_s$ with smaller resolution (112$\times$112 in this work) to explicitly learn the consistent representations among different scales. It is worth to note that the additional view with smaller scale only consumes marginal cost, keeping the efficiency of the original pipeline. 
	
\begin{figure}[t]
		\begin{center}
			\includegraphics[width=1.03\linewidth]{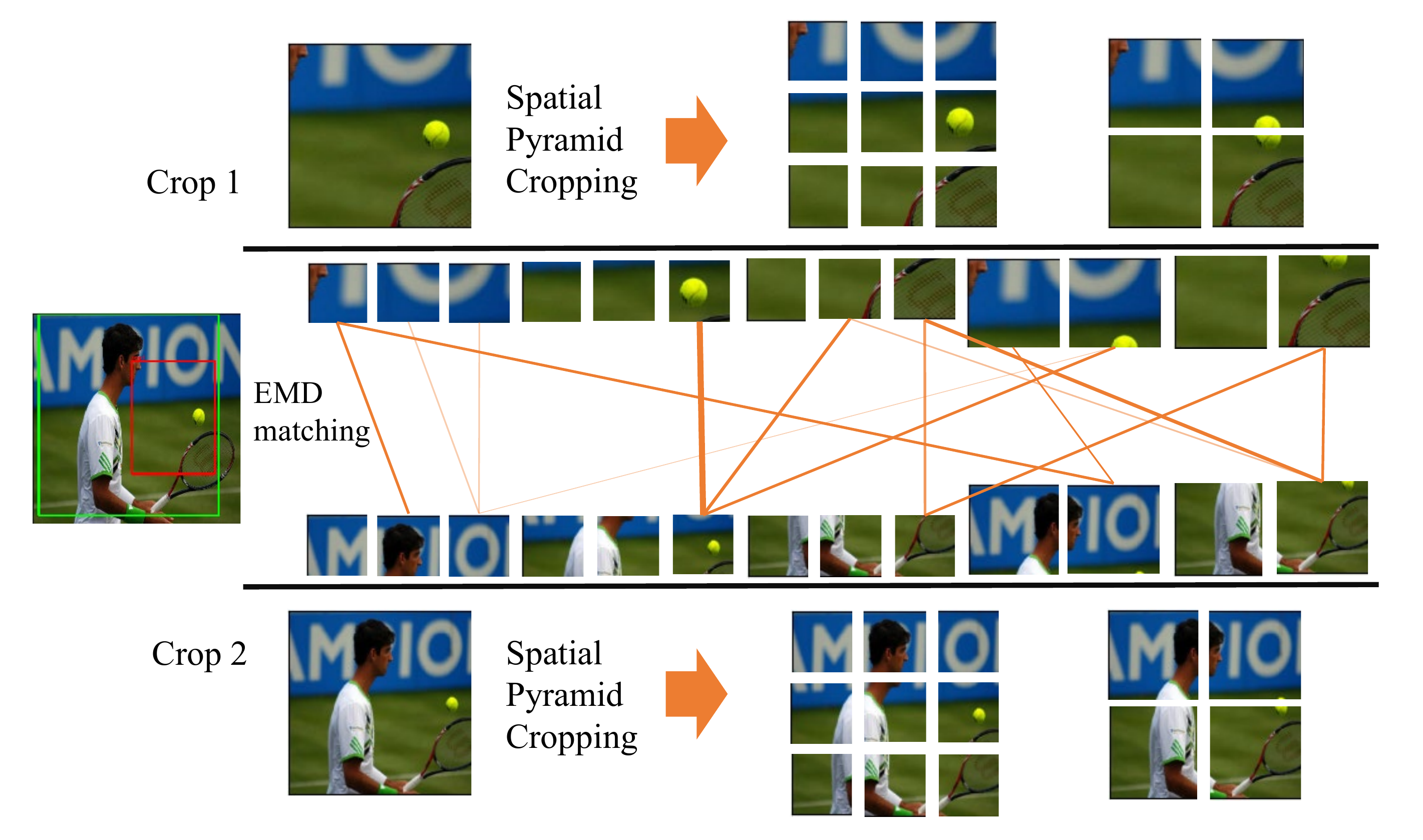}
		\end{center}
		\caption{Spatial pyramid crops for \textit{Self}-EMD. We parallelly conduct several average pooling layers with different kernel sizes and strides to generate the spatial pyramid crops. We project the crops of feature maps back to the image for better visualization.}
		\label{fig:emd}
	\end{figure}

	Furthermore, inspired by Spatial Pyramid Pooling (SPP)~\cite{spp}, we resize the feature representations into different grid sizes (\emph{i.e.}, $7\times7$, $5\times5$, and $3\times3$) via several average pooling layers with different kernel sizes and strides to generate the set of Spatial Pyramid Cropping (SPC) for EMD, as depicted in Fig.~\ref{fig:emd}. This operation makes the comparison of local features can across different scales from the two different crops, enhancing scale information in the learned local representations.
	
\section{Experiments}
	We first present implementation details in our training framework. Then we compare our method and other methods on several detection benchmarks (\emph{e.g.}, COCO~\cite{MSCOCO}, Pascal VOC~\cite{Pascal-voc}) with some typical detection models.    
	
	\begin{table*}[!thbp]
		\begin{center}
			\scalebox{0.95}{
				\resizebox{\textwidth}{!}{
					\begin{threeparttable}
						\begin{tabular}{l|l|c|c|c|c|c|c|c|c}
							\hline
							Detector                   & Pre-train Method & \begin{tabular}[c]{@{}c@{}}Pre-train\\ Data\end{tabular} & \begin{tabular}[c]{@{}c@{}}Pre-train\\ Label\end{tabular} & $AP$   & $AP_{50}$ & $AP_{75}$ & $AP_s$  & $AP_m$  & $AP_l$  \\ \hline
							\multirow{7}{*}{FPN}   & Classification   & ImageNet                                                 &      \checkmark                                         & 39.1 & 60.0 & 42.2 & 24.1   & 42.7    & 50.4   \\
							& BYOL             & ImageNet                                                 &                                                           & 39.9 & 60.2 & 43.2  &  23.3 & 43.2 & 52.8 \\
							& \textit{Self}-EMD              & ImageNet                                                 &                                             & 40.0  & 60.4 & 44.0  &  23.5 & 43.8 & 52.2 \\ \cline{2-10} 
							& BYOL             & COCO                                                     &                                                           & 38.8 &  58.5 & 42.2  &  23.3 &  41.4 & 49.5  \\
							& \textit{Self}-EMD              & COCO                                                     &                                             & \textbf{39.8}  &  60.0 & 43.4 & 24.2  & 42.7 & 50.6  \\ \cline{2-10}
							& BYOL             & COCO+                                                     &                                                           & 39.3 &  59.0 & 42.8  &  23.5 &  42.1 & 50.5  \\
							& \textit{Self}-EMD              & COCO+                                                     &                                             & \textbf{40.4}  &  61.1 & 43.7 & 24.4  & 43.3 & 51.3  \\ \hline
							\multirow{3}{*}{Mask+C4}  & Classification   & ImageNet                                                 &       \checkmark                                          & 38.2 & 58.2 & 41.2 & 21.6 & 42.7 & 52.1 \\
							& BYOL             & COCO                                                     &                                                           & 37.9 & 57.5 & 40.9  &  21.6 &  42.6  &  51.2 \\
							& \textit{Self}-EMD              & COCO                                                     &                                                           & \textbf{38.5} & 58.3 & 41.6 &  21.5 & 43.3  & 51.9  \\ \hline
							\multirow{3}{*}{Mask+FPN}       & Classification   & ImageNet                                                 &      \checkmark                                           & 38.9 & 59.6 & 42.7 & 23.7 & 42.6 & 52.0 \\
							& BYOL             & COCO                                                     &                                                           & 38.5 & 58.9 & 41.7 &  22.8 & 41.9 & 49.3 \\
							& \textit{Self}-EMD              & COCO                                                     &                                                           & \textbf{39.3} & 60.1 & 42.8  & 24.4 & 42.7 & 49.9 \\ \hline
							\multirow{3}{*}{RetinaNet} & Classification   & ImageNet                                                 &     \checkmark                                             & 37.1 & 56.1  & 39.7 & 22.9  & 41.1  & 48.4  \\
							& BYOL             & COCO                                                     &                                                           & 36.2 & 54.8 & 38.8 & 21.3 & 40.2 &  45.4 \\
							& \textit{Self}-EMD              & COCO                                                     &                                                           & \textbf{37.4}  & 56.5 & 39.7 & 23.2 & 41.3 & 46.6  \\ \hline
						\end{tabular}
			\end{threeparttable}}}
		\end{center}
		\caption{Main detection performance of several typical detectors in terms of AP (\%) on COCO \emph{val} with standard 1$\times$ schedule. `COCO+' denotes the COCO \textit{train} 2017 set plus the COCO \textit{unlabel} set. For fair comparison, BYOL~\cite{byol} and \textit{Self}-EMD are both pre-trained for 300 epochs on ImageNet and 800 epochs on COCO.}
		\label{table:main}
	\end{table*}
	
	\subsection{Implementation details}
	\paragraph{Data.} For training data, we consider using a cluttered dataset against the object-centered ImageNet dataset to evaluate our experiments. Unless otherwise specified, we pre-train the network on the COCO 2017 \textit{train} set without labels. For detection fine-tuning, we follow the official splits to train and evaluate the detection model on COCO and Pascal VOC benchmarks.   
	
	\paragraph{Image augmentations.} We use the same set of image augmentations as in BYOL. A random patch of the image is cropped and resized to $224 \times 224$, followed by a color distortion, a random Gaussian blur, a random horizontal flip and a solarization. For the additional view $v_s$, the corresponding patch is resized to $112 \times 112$. For detection fine-tuning, we follow the common practice of Detectron2~\cite{detectron2} to adopt random horizontal flip and multi scale training. 
	
	\paragraph{Architecture.} Following the common practice, we use a standard ResNet-50~\cite{ResNet} as our base backbone in two encoders. As mentioned in Sec.~\ref{sec:3.2}, we remove the global pooling layer and replace the linear layer in the MLP head with $1\times1$ convolution layer. To calculate the marginal weights in Eq.~(\ref{margin:weights}), we still keep the original MLP head as a parallel branch to generate vector features.   
	
	
	\paragraph{Optimization.} For pre-training on COCO, we use the SGD optimizer with a 1e-4 weight decay. To keep the total training iterations comparable with the ImageNet supervised training, we train the models on COCO for 800 epochs on 8 NVIDIA 2080ti GPUs with the total batch size of 256. Warm-up is used at the first 4 epochs, where the learning rate starts from 0.0 and then linearly increases to 0.2. Afterwards, the learning rate decreases to 0.0 following a cosine scheduler.

	\subsection{Main Results}
	We conduct the proposed \textit{Self}-EMD and BYOL pre-training on both the ImageNet train set and the COCO \textit{train} 2017 set, and then fine-tune several typical detection models on COCO dataset for 90k iterations (standard 1$\times$ schedule in Detectron2~\cite{detectron2}). Note that unsupervised pre-training can have different distribution and statistics compared with ImageNet supervised pre-training, so we follow the practice in \cite{moco} that fine-tune with Synchronized BN (SyncBN) and add SyncBN to newly initialized layers (\emph{e.g.}, FPN~\cite{FPN}). The image scale is [640, 800] pixels during training and 800 at inference. For fair comparison, we also employ the same strategy to fine-tune the random initialized weights and the supervised pre-training baseline.
	
	As reported in Table~\ref{table:main}, on Faster R-CNN~\cite{FasterCNN} with FPN, when pre-training on ImageNet, \textit{Self}-EMD and BYOL get 40.0\% mAP and 39.9\% mAP respectively, both surpassing the performance of the supervised pre-training model (39.1\% mAP). However, when pre-training on the non-iconic dataset COCO, the AP of BYOL decreases by 1.1\% (from 39.9\% to 38.8\%), indicating that the cluttered noise is indeed a problem for self-supervised training.
	
	In contrast, our \textit{Self}-EMD model trained on COCO can achieve comparable results (39.8\% mAP) with various models (\emph{i.e.}, supervised, BYOL, \textit{Self}-EMD) trained on ImageNet. These results demonstrate the effectiveness of out method for addressing the non-iconic issue. We further conduct additional experiments on other detectors compared to BYOL. For Mask R-CNN with FPN, the AP increases from 38.5\% to 39.3 \% (+0.8\%), for Mask R-CNN with C4 architecture, the AP increases from 37.9\% to 38.5\% (+0.6\%). We also investigate the typical single stage model RetinaNet, and the AP increases from 36.2\% to 37.4\% (+1.2\%).      
	
	
	
	
	We notice that the improvement is most significant in the single stage detector RetinaNet. One possible reason is that for single stage detector, the local feature representations are more important as the prediction is directly from the convolutional feature maps of the backbone. In other words, as EMD strategy preserves the spatial structures and local information during the pre-training process, the learned representations are more suitable for dense prediction.          
	
	To further explore the potential of the proposed \textit{Self}-EMD, we conduct an experiment with more training data on the standard Faster Res50-FPN detector. Specifically, we add the additional COCO 2017 \textit{unlabel} set which contains 123k unlabeled images for \textit{Self}-EMD pre-training. Results on Fig.~\ref{fig:0} and Table~\ref{table:main} show that our method achieves further improvement with more data, which demonstrate the superiority of self-supervised pre-training. 
	
	\subsection{Ablation Study}
	In this section we conduct several ablation studies to verify the proposed EMD strategy. All experiments use standard Faster R-CNN Res50-FPN detector to evaluate the detection performance on COCO 2017 val set. 
	
	\paragraph{Marginal Weights} We first investigate the marginal weights designing, which is the crucial module for applying optimal transport model. As shown in Table~\ref{table:ablation1}, when we follow \cite{deepemd} to adopt Eq. (\ref{eq:weights}) in the 3rd row, the resulting detection performance decreases from 38.8\% mAP to 37.2\% compared to the original BYOL pre-training. As we analyzed in Section~\ref{sec:3.2}, the working mechanism of EMD highly depends on the precondition that the convolutional feature maps have informative semantic meanings. Although such condition is naturally satisfied under supervised training, it is non-trivial when training the network without labels. Further, when we replace the marginal weights with Eq. (\ref{margin:weights}), the performance is improved by 2.3\% (from 37.2\% mAP to 39.5\%). This improvement verifies the effectiveness of the proposed EMD metric, and it also highlights the importance of the marginal weights when applying EMD metric for self-supervised learning.           
	
	\begin{table}[hbp]
		\begin{center}
			\scalebox{0.47}{
				\resizebox{\textwidth}{!}{
					\begin{threeparttable}
						\begin{tabular}{l|cc|lll}
							\hline
							\begin{tabular}[c]{@{}l@{}}Pre-train\\ Methods\end{tabular}     & \begin{tabular}[c]{@{}l@{}}Marginal\\ Weights\end{tabular} &  \begin{tabular}[c]{@{}l@{}}Scale\\ Training\end{tabular} & $AP$   & $AP_{50}$ & $AP_{75}$ \\ \hline
							Random init.                 &       --                                                    &                --                                          & 34.7 & 52.9  &  35.3 \\
							BYOL                 &             --                                                  &                    --                                      & 38.8 & 58.5 & 42.2  \\ \hline
							\multirow{3}{*}{\textit{Self}-EMD} &         Eq.~\ref{eq:weights}                                          &                                                          & 37.2 & 55.7 & 40.6  \\
							& Eq.~\ref{margin:weights}                                      &                                                          & 39.5 & 59.6  &  43.0 \\
							& Eq.~\ref{margin:weights}                                         & \checkmark              & 39.8 & 60.0 &  43.4 \\ \hline
						\end{tabular}
			\end{threeparttable}}}
		\end{center}
		\caption{Detection performance in terms of AP (\%) on COCO \emph{val}. All methods are conducted with standard Faster Res50-FPN detector.}
		\label{table:ablation1}
	\end{table}
	
	\paragraph{Scale Invariant Training} In Table~\ref{table:ablation1}, we also investigate the advanced tricks of scale invariant training as depicted in Section~~\ref{sec:3.3}. Specifically, the proposed Spatial Pyramid Crops (SPC) and the multi scale training further boost the AP by 0.3\% (from 39.5\% to 39.8\%). Although the improvement is not significant, this consistent gain indicates the importance of scale invariance in object detection. Furthermore, we suspect the possible reason of minor improvement is that the strong data augmentation of self-supervised method has already introduced scale-invariant representations in some degree. We speculate that designing a better data augmentation strategy for object detection may bring more robust scale representations, and we will leave this exploration for future work.         
	
	\begin{table}[thbp]
		\begin{center}
			\scalebox{0.47}{
				\resizebox{\textwidth}{!}{
					\begin{threeparttable}
						\begin{tabular}{l|l|c|c|c|c}
							\hline
							Backbone                     & Pre-train Method & \begin{tabular}[c]{@{}c@{}}Pre-train\\ Data\end{tabular} & $AP$   & $AP_{50}$ & $AP_{75}$  \\ \hline
							\multirow{3}{*}{MobileNetV2}  & Classification   & ImageNet                                                 & 29.3 & 47.9 & 30.9   \\
							& BYOL             & COCO                                                                                  & 26.0 & 43.4  & 27.1  \\
							& \textit{Self}-EMD              & COCO                                                                    & 27.1 & 44.9  & 28.4  \\ \hline
							\multirow{3}{*}{ResNet 101} & Classification   & ImageNet                                                  & 40.2 & 61.0 & 43.0  \\
							& BYOL             & COCO                                                                                 & 41.0 & 61.0 & 45.0  \\
							& \textit{Self}-EMD              & COCO                                                                    & \textbf{41.6} & 61.9 & 45.5   \\ \hline
						\end{tabular}
			\end{threeparttable}}}
		\end{center}
		\caption{Detection performance with different backbones in terms of AP (\%) on COCO \emph{val}.}
		\label{table:backbone}
	\end{table}
	\paragraph{Training efficiency} For the implement efficiency of EMD loss, we show the comparison of training time and speed (Sec./Iter) in Table~\ref{table:speed}. As we adopt the efficient Sinkhorn Iteration (we fix the iter at 10) to calculate EMD loss, the additional training time is small.
	
	\begin{table}[!h]
		\begin{center}
			\scalebox{0.45}{
				\resizebox{\textwidth}{!}{
					\begin{threeparttable}
						\begin{tabular}{l|l|l}
							\hline
							Methods (800 epochs on COCO) & Total time & Sec./Iter \\ \hline
							BYOL                         & 39h        & 0.38s     \\
							Self-EMD w/o SPC             & 44h        & 0.43s     \\
							Self-EMD w/ SPC              & 45h        & 0.44s     \\
							Self-EMD w/ scale training   & 52h        & 0.51s     \\ \hline
						\end{tabular}
			\end{threeparttable}}}
		\end{center}
		\caption{Comparison of training time.}
		\label{table:speed}
	\end{table}  
	
	\paragraph{Different Backbone Structures} We implement \textit{Self}-EMD on different backbone structures to evaluate the versatility in Table~\ref{table:backbone}. For small model (\emph{i.e.}, MobileNetV2~\cite{mobilenetv2}), although consistent boost is achieved by \textit{Self}-EMD, the overall performance of self-supervised models are inferior to the supervised pre-training. While for large model (\emph{i.e.}, ResNet 101), the advantage of self-supervised pre-training over the supervised model extends.   
	
	
	\subsection{Compare to other methods}
	In this section, we compare our method with other self-supervised methods, \emph{e.g.}, SimCLR~\cite{simclr} and MoCov2~\cite{mocov2}, as well as other supervised pre-training method. All experiments adopt standard Faster Res50-FPN detector to evaluate the detection performance on COCO 2017 val set. 
	
	As shown in Table~\ref{table:others}, when pre-training on COCO, our proposed method outperforms the other self-supervised methods by a larger margin. The potential reason is that for SimCLR and MoCov2, they need additional negative samples in the contrastive learning manner, which may intensify the reliance of object-centric bias. When we look at the shrinkage from pre-training on ImageNet to COCO, SimCLR and MOCO decreases by 1.8\% (from 39.4\% to 37.6\%) and 1.4\% (from 39.6\% to 38.2\%) respectively, where the performance change of our \textit{Self}-EMD is much stable (from 39.6\% to 39.8\%). 
	
	We also achieve comparable result with other supervised pre-training method. Montage~\cite{cheaper} is a recent detection pre-training method that only use the label of detection dataset to crop the image and then ensemble the cropped samples in a Montage manner for classification pre-training. For fair comparison, we also add Sync BN for training the detector with the pre-trained weights from our re-implementation. The results show that our method significantly surpass the supervised pre-training method even without any label.   
	
	\begin{table}[htbp]
		\begin{center}
			\scalebox{0.47}{
				\resizebox{\textwidth}{!}{
					\begin{threeparttable}
						\begin{tabular}{l|c|c|c|c|c}
							\hline
							Pre-train Method & \begin{tabular}[c]{@{}c@{}}Pre-train\\ Data\end{tabular} & \begin{tabular}[c]{@{}c@{}}Pre-train\\ Label\end{tabular} &$AP$ & $AP_{50}$ &$AP_{75}$\\ \hline
							Classification   & ImageNet                                                 &  \checkmark                                               & 39.1 & 60.0 & 42.2 \\
							Montage          & COCO                                                     &  \checkmark                                               & 39.0 & 60.2 & 41.0 \\ \hline
							MOCOv2             & ImageNet                                                 &                                                           & 39.6 & 59.5 & 43.2 \\
							SimCLR           & ImageNet                                                 &                                                           & 39.4 & 59.1 & 42.9 \\ 
							BYOL             & ImageNet                                                 &                                                           & 39.9 & 60.2 & 43.2 \\
							\textit{Self}-EMD& ImageNet                                                 &                                                           & 40.0 & 60.4 & 44.0 \\\hline
							MOCOv2             & COCO                                                     &                                                           & 38.2 & 58.9 & 41.6 \\
							SimCLR           & COCO                                                     &                                                           & 37.6 & 58.3 & 40.3 \\
							BYOL             & COCO                                                     &                                                           & 38.8 & 58.5 & 42.2 \\
							\textit{Self}-EMD& COCO                                                     &                                                           & \textbf{39.8} & 60.0 & 43.4 \\ \hline
							
						\end{tabular}
			\end{threeparttable}}}
		\end{center}
		\caption{Detection performance comparison with other pre-training methods in terms of AP (\%) on COCO \emph{val}.}
		\label{table:others}
	\end{table} 
	
	\subsection{Results on Pascal VOC}
	\begin{table}[!h]
		\begin{center}
			\scalebox{0.45}{
				\resizebox{\textwidth}{!}{
					\begin{threeparttable}
						\begin{tabular}{l|c|c|c|c}
							\hline
							Pre-train Method & \begin{tabular}[c]{@{}c@{}}Pre-train\\ Data\end{tabular} & $AP$ &$AP_{50}$&$AP_{75}$\\ \hline
							Classification   & ImageNet (sup.)                                                & 53.4 & 80.1 & 59.4  \\ \hline
							Detection        & COCO (sup.)                                                    & 56.8 & \textbf{83.2} & 63.7  \\ \hline
							MOCOv2             & COCO                                                     & 51.9 & 78.9 & 58.0 \\
							BYOL             & COCO                                                     & 50.0 & 75.8 & 54.8\\
							\textit{Self}-EMD& COCO                                                     & 53.0 & \textbf{80.0} & 58.6  \\ \hline
						\end{tabular}
			\end{threeparttable}}}
		\end{center}
		\caption{Detection performance with several pre-training methods in terms of AP (\%) on VOC 2007 \emph{test}.}
		\label{table:voc}
	\end{table} 
	We examine the proposed pre-training strategy on Pascal VOC dataset with the standard Faster Res50-FPN detector. We fine tune several typical detectors on the VOC trainval 07+12 set and evaluate the performance on the VOC test 2007 set, following the common practice. Similar to the settings of COCO, we adopt SyncBN and multi scale training. For evaluation, we conduct the default VOC metric of $AP_{50}$ (\emph{i.e.}, IoU threshold is 50\%) and the COCO-style metric. The results in Table~\ref{table:voc} show that our proposed method is the best among all self-supervised methods with a large margin. Compared with BYOL, \textit{self}-EMD outperforms it by 3.0\% and 4.2\% on AP and $AP_{50}$, respectively. 
	
	However, we also notice there is a significant gap for all the self-supervised methods compared with supervised detection pre-training. This result indicates that the benefit from current self-supervised methods is limited. Designing a better self-supervised training framework is still an open question for object detection pre-training 
	
	
	\subsection{Visualization}
	 	
	\begin{figure}[t]
		\begin{center}
			\includegraphics[width=0.999\linewidth]{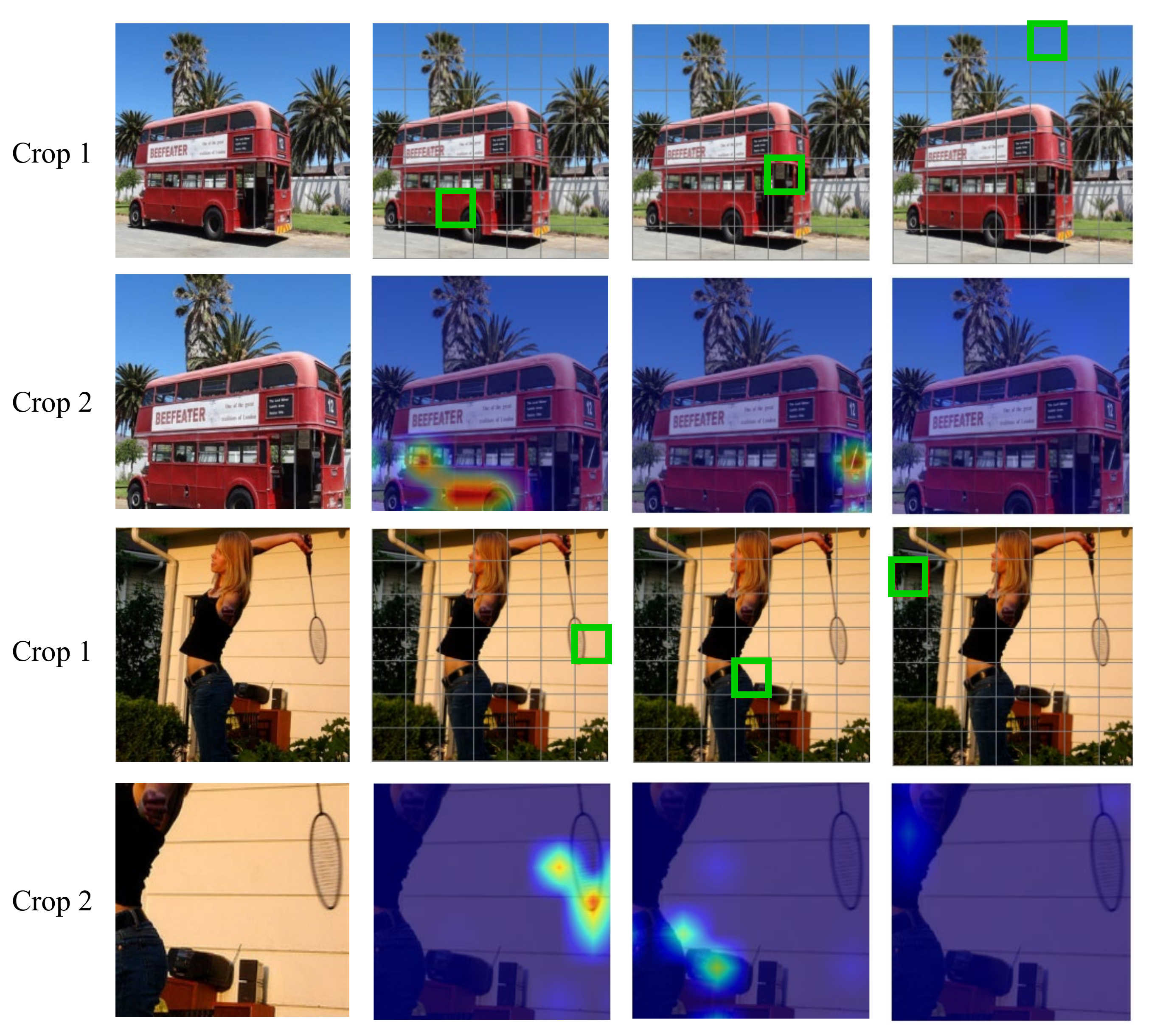}
		\end{center}
		\caption{Visualization of the optimal matching weights. We visualize the heat map of the matching weights of the crop 2 to the corresponding position (the green boxes) in the crop 1.}
		\label{fig:vis}
		\vspace{-0.2cm}
	\end{figure}
	
	We visualize the optimal matching weights of \textit{Self}-EMD with the pre-trained network. We only adopt EMD on the original $7\times7$ feature maps without spatial pyramids cropping. In Fig.~\ref{fig:vis}, we first generate two random crops from the same image where the color jitters are not shown for better visualization. Then we visualize heat map of the weights of crop 2 to the corresponding position in the crop 1. As we can see, \textit{Self}-EMD can effectively establish semantic correspondence between two crops, and the background in the crop 1 or the irrelevant regions in the crop 2 are assigned with small weights, thus alleviating the cluttered noise during training. 
	
\section{Conclusion}
	In this paper, we have presented a novel self supervised method named \textit{Self}-EMD to learning spatial visual representations for object detection. It keeps the convolutional feature maps as the image embedding and then employs a discrete Earth Mover's Distance to measure the spatial similarity. 
	\textit{Self}-EMD can achieve the leading result even without ImageNet dataset. This enables us to exploiting more unlabeled data in the future.

	{\small
		\bibliographystyle{ieee_fullname}
		\bibliography{egbib}
	}
	
\end{document}